\documentclass[10pt]{article}
\usepackage{fullpage}
\usepackage{amsmath}
\usepackage[amsthm, thmmarks]{ntheorem}
\usepackage{amssymb}
\usepackage{graphicx}
\usepackage{enumerate}
\usepackage{verse}
\usepackage{tikz}
\usepackage{verbatim}
\usepackage{hyperref}
\usepackage{algorithm}
\usepackage{clrscode3e}
\usepackage{xcolor}
\usepackage{adjustbox}
\usepackage{multirow}
\usepackage{wrapfig}

\newcommand{\ve}[1]{\mathbf{#1}}

\newcommand{\mrm}[1]{\mathrm{#1}}
\newcommand{\etal}{{et~al.}}

\newcommand{\mcal}[1]{\mathcal{#1}}

\title{Talking Head(?) Anime from a Single Image 4:\\ Improved Model and Its Distillation}
\author{Pramook Khungurn\\pixiv Inc.}

\begin{document}
\maketitle

\begin{abstract}
  We study the problem of creating a character model that can be controlled in real time from a single image of an anime character. A solution to this problem would greatly reduce the cost of creating avatars, computer games, and other interactive applications. 
  
  Talking Head Anime 3 (THA3) is an open source project that attempts to directly address the problem \cite{Khungurn:2022:THA3}. It takes as input (1) an image of an anime character's upper body and (2) a $45$-dimensional pose vector and outputs a new image of the same character taking the specified pose. The range of possible movements is expressive enough for personal avatars and certain types of game characters. However, the system is too slow to generate animations in real time on common PCs, and its image quality can be improved.

  In this paper, we improve THA3 in two ways. First, we propose new architectures for constituent networks that rotate the character's head and body based on U-Nets with attention \cite{Ho:2020} that are widely used in modern generative models. The new architectures consistently yield better image quality than the THA3 baseline. Nevertheless, they also make the whole system much slower: it takes up to 150 milliseconds to generate a frame. Second, we propose a technique to distill the system into a small network ($< 2$ MB) that can generate $512\times512$ animation frames in real time ($\geq 30$ FPS) using consumer gaming GPUs while keeping the image quality close to that of the full system. This improvement makes the whole system practical for real-time applications.
\end{abstract}

\section{Introduction}

We are interested in animating a single image of an anime character through specifying explicit pose parameters, as if controlling a rigged 3D model. Our motivation is the recent popularity of \emph{virtual YouTubers} (VTubers), which are anime characters that are performed in real time by actors with the help of recent computer graphics technologies \cite{Lufkin:2018}. Typically, VTubers employ controllable layered images (aka 2.5D models) \cite{Litwinowicz:1991} created by software such as Live2D \cite{Live2D:2023}, E-mote \cite{Emote:2023}, and Spine \cite{Spine:2023}. Because such a model can be costly to create, a solution would make it much easier to acquire a controllable avatar and to produce computer games and other interactive media.

The problem has received some attention from the research community \cite{Zhang:2020, Kim:2021:AnimeCeleb}, private companies \cite{Iriam:2021, Algoage:2022}, and individual open-source developers \cite{Khungurn:2022:THA3, Transpchan:2022}. In particular, Khungurn proposes a neural network system called ``Talking Head Anime 3'' (THA3) that can generate simple animations of a humanoid anime character, given only a single image of the frontal view of the character's torso \cite{Khungurn:2022:THA3}. When the system is run on a powerful GPU, the character can be controlled interactively through 45 parameters, enabling rich facial expressions and rotation of the head and the body by small angles. With NO manual modeling, the system can generate movements that are similar to what typical hand-made VTuber models are capable of.

Nevertheless, the THA3 system has two main limitations. The first is the quality of the generated images. When occluded parts of the character are rotated and become visible, they are often blurry. Moreover, the system has a tendency to remove thin structures, such as hair stands after head rotation. The second is the system's speed: interactive frame rates can only be achieved when using very powerful GPUs, such as the Nvidia Titan RTX \cite{Khungurn:2022:THA3}. As a result, it is not yet a practical system for real-time applications such as computer games or VTuber streaming. 

This paper proposes two improvements to the THA3 system to address the two shortcomings above. The first improvement is a new architecture for the subnetworks that rotate body parts. THA3 uses an encoder-decoder network and a vanilla U-Net as described in the original paper \cite{Ronneberger:2015}. Our new architecture is based on the variant of U-Net with attention \cite{Vaswani:2017}, commonly used in diffusion probabilistic models \cite{Ho:2020, Dhariwal:2021}. It improves image quality under three commonly used metrics, reduces blurring in disoccluded\footnote{``Disocclude'' means ``to cause to be no longer occluded.'' As far as we know, the word does not appear in standard dictionaries but has been used in a number of computer vision papers \cite{Park:2017,Yang:2015}.} areas, and preserves thin structures better. Unfortunately, it sacrifices inference speed to do so, taking more than $100$ ms to generate a frame on the Titan RTX.

The second improvement directly addresses the speed problem. The idea is to distill \cite{Hinton:2015} the knowledge of the full system into a student neural network is small ($\approx 2$ MB) and can generate a $512 \times 512$ frames in no more than $30$ ms using a consumer gaming GPU. However, the student is specialized to a specific input image and cannot animate any other. For a given character image, distillation takes several tens of hours, but, once the process is finished, the student network can be used as a controllable character model. This capability makes the THA system usable in real-time applications for the first time. While we do lose the ability to change character and animate it immediately, the system remains practical because a VTuber or a game character does not change its appearance so often (every second or every minute).

The architecture for the student network is based on the SInusoidal REpresentation Network (SIREN) \cite{Sitzmann:2020}, which we extend to make it faster and better at preserving details of the input image. To make SIREN faster, we make it generate images in a multi-resolution fashion similar to the way a GAN generates an image \cite{Radford:2016}: the first few layers generate a low-resolution feature tensor, which is upscaled and passed to later layers. To better preserve the details of the input image, we have the SIREN generate appearance flow \cite{Zhou:2017} that is used to warp the input image. The result is then combined through alpha blending with another less detailed output, also produced directly by SIREN. We demonstrate through ablation studies that our proposed architecture achieves a good trade-off between speed and accuracy, being able to reproduce high-frequency details while generating large images in real time. We also propose a three-phase training process that is effective at training our proposed model, and we verify that all of the phases are necessary to achieve better image quality.

\section{Related Works} \label{sec:related-works}

\subsection{Implicit Neural Representation}

The student network is a special case of {\bf neural implicit representations} (INRs) where neural networks are used to approximate signals rather than functions that transform them. INRs often incorporate positional encoding \cite{Tancik:2020} or have unconventional activation functions \cite{Sitzmann:2020, Saragadam:2023} or network structures \cite{Singh:2023}. Researchers have applied INRs to signals such as images \cite{Stanley:2007, Chen:2021}, 3D surfaces \cite{Mescheder:2019, Park:2019}, and volume density coupled with radiance \cite{Mildenhall:2020}. INRs can be used to build generative models of high-resolution 3D signals, which were hard to achieve previously \cite{Chen:2019, Chan:2021, Chan:2022, Schwarz:2021, Sun:2023}. 

While INRs can be used to directly represent articulated characters \cite{Deng:2020,Yenamandra:2021,Peng:2021,Zhuang:2022}, we take the view that our signal is a parameterized collection of images like Bemana \etal~\cite{Bemana:2020} rather than a deformable 3D shape. As a result, our INR employs the same image processing techniques such as warping and interpolation. However, our work is different from Bemana \etal's in two ways. Firstly, the parameters of our image collection are blendshape weights and joint angles rather than those related to viewpoint, lighting, and time. Secondly, we use an INR to approximate the outputs of a bigger neural network rather than to fill in the gap between sparse measurements.

\subsection{Parameter-Based Posing of a Single Image}

Overall, we want to create simple animations from a single image of a humanoid character. The input is an image of a subject (the {\bf target} image), and we need to modify it so that the subject is posed according to some specification. Based on how pose is specified, the problem can be classified into {\bf parameter-based posing} (explicitly by a {\bf pose vector}), {\bf motion transfer} (implicitly via an image or video of another subject), or {\bf visual dubbing} (inferred from a spoken voice record). Our system solves parameter-based posing of a single image. As a result, we will review works that solve the same problem, excluding those that take videos or multiple images as input. To our knowledge, there are three approaches to the problem at hand.

\subsubsection{Direct Modeling}

We can create a controllable model of the subject's geometry from the target image. The common approach is to fit a {\bf 3D morphable model} (3DMM) \cite{Blanz:1999, Cao:2014, Loper:2015, Li:2017, Pavlakos:2019, Osman:2020, Zheng:2022} to the image. While earlier works are limited in controllability and only suitable for image manipulation \cite{Blanz:2004, Cao:2014, Fried:2006}, recent works provide much more control \cite{Geng:2019, Hong:2022, Lin:2023, He:2021:Arch, Corona:2023, Lattas:2023}. A limitation of this approach is that parametric models often do not model all visible parts. For example, models specialized to the face might ignore the hair \cite{Lin:2023, Lattas:2023, Lattas:2023}, the neck \cite{Hong:2022}, or both \cite{Geng:2019}.

While much research has been done on modeling from human photos, much less attention has been paid to other image domains. Saragih \etal\ construct a 3D model of a non-human face and then deform it to create animations \cite{Saragih:2011}, but they can only animate masks. Jin creates E-mode models from single anime-style images \cite{Jin:2020}. Chen \etal\ study 3D reconstruction from a single anime character's image \cite{Chen:2023:CVPR} where the reconstruction can be later animated with the help of off-the-shelf components \cite{Xu:2020,Khungurn:2021:THA2}. 

\subsubsection{Generative Modeling in the Latent Space}

We first train a generative model that maps a {\bf latent code} to an image, engineering it so that the output image is controllable through a pose vector. At test time, we fit a latent code to the target image.\footnote{Optionally, the generative model can be fine-tuned to match the input image better \cite{Roich:2021}.} Animation frames can then be generated by fixing the latent code and varying the pose vector.

Much research has been done on controlling the human face. Tewari \etal\ trains a network to alter latent codes of a StyleGAN \cite{Karras:2019, Karras:2020} according to 3DMM parameters \cite{Tewari:2020:StyleRig} and later proposes a specialized algorithm to fit latent codes to portraits \cite{Tewari:2020:PIE}. Using different methods, Kowalski \etal\ \cite{Kowalski:2020} and Deng \etal\ \cite{Deng:2020:DiscoFaceGan} train GANs, each of whose latent codes have parts that are explicitly controllable. Recent works extend EG3D \cite{Chan:2022}, a 3D-aware GAN, so that the facial expression can be controlled \cite{Ma:2023, Xu:2023:OmniAvatar, Sun:2023} 

\subsubsection{Image Translation}

Alternatively, we can view parameter-based posing as a special case of {\bf image translation}: transforming an image into another according to some criteria. Isola \etal\ \cite{Isola:2017} present a general framework based on conditional generative adversarial networks (cGANs) \cite{Mirza:2014}, which is extended in various aspects by subsequent research \cite{Zhu:2017, Choi:2018, Choi:2020}. Recently, researchers have also started exploring using diffusion models for the task \cite{Saharia2022, Wu:2022, Li:2023}.

Pumarola \etal\ create a network that modifies human facial features given an \emph{Action Units} (AUs) encoding of a facial expression \cite{Pumarola:2019}. Ververas and Zafeiriou do the same but use blendshape weights instead of the AUs \cite{Ververas:2020}. Ren \etal's PIRenderer handles not only facial expression but also head rotation \cite{Ren:2021}. Zhang \etal's SadTalker \cite{Zhang:2023} can control a face image through 3DMM parameters by mapping them to facial landmark positions, which are then fed to Chen \etal's face-vid2vid model \cite{Wang:2021} to move the input image. Nagano \etal\ design a conditional GAN that outputs a realistic facial texture, taking as input the target image and renderings of a template mesh whose expression can be freely controlled \cite{Nagano:2018}.

The THA3 system \cite{Khungurn:2022:THA3} that we build upon is based on Pumarola \etal's facial feature manipulation technique and Zhou \etal's image rotation technique \cite{Zhou:2016}. Zhang \etal\ extended the first version of THA \cite{Khungurn:2019:THA} in order to support larger rotation angles \cite{Zhang:2020}. Kim \etal\ created a dataset that can be used to train parameter-based posers such as PIRenderer so that they work on anime characters \cite{Kim:2021:AnimeCeleb}.

\section{Baseline} \label{sec:baseline}

THA3 as a whole is an image translator that takes as inputs (1) an $512 \times 512$ image of the ``half body shot'' of a humanoid anime character and (2) a 45-dimensional pose vector and then outputs a new image of the same character, now posed accordingly. The 45 parameters allow a character to not only express various emotions but also move its head and body like a typical professionally-created VTuber model. Out of the 45 parameters, 39 control the character's facial expressions, and 6 control rotation of the face and the torso.

The system is composed of 5 neural networks, and they can be divided into two modules. Three networks form a module called the {\bf face morpher} whose duty is to alter the character's facial expression. We will not modify this module, but we will distill it into a smaller network in Section~\ref{sec:distillation}. The remaining two networks are called the {\bf half-resolution rotator} and the {\bf editor}. Together, they form a module called the {\bf body rotator} whose duty is to rotate the head and the torso according to the 6 non-facial-expression parameters. The half-resolution rotator operates on a half-resolution ($256 \times 256$) image obtained by downscaling the output of the face morpher. Its output is then upscaled to $512\times512$ and then passed to the editor to improve image quality before finally being returned to the user.

The two networks share the same overall structure. Each contains a backbone convolutional neural network (CNN): the half-resolution rotator uses an encoder-decoder network, and the editor uses a U-Net. Each backbone network outputs a feature tensor that has the same resolution as the input image. The feature tensor can then be used to perform three image processing operations:
\begin{enumerate}
  \item {\bf Warping.} The feature is transformed into an \emph{appearance flow}, a map that tells for each pixel in the output which pixel in the input should data be copied from \cite{Zhou:2017}. The appearance flow is then applied to the input image to get a warped version of it.
  \item {\bf Direct generation.} The feature tensor is directly transformed into pixel values. This operation is not limited by what is visible in the input image. It yields more plausible disoccluded parts but cannot preserve all the details in the visible parts.
  \item {\bf Blending.} The feature tensor is transformed into an alpha map, which can then be used to blend the results of other steps together to get the best features of both operations.
\end{enumerate}
Outputs of the networks are generated using some combinations of the above operations.

\section{Improved Network Architecture} \label{sec:new-architecture}

One of the main problems of THA3 is image quality. When the body parts are rotated and disoccluded parts become visible, these parts can be blurry. Moreover, if such parts are thin, the system tend to remove them altogether. To alleviate the problem, we modify the networks in the body rotator module without significantly changing their functions. 

\subsection{New Body Rotator Architecture}

\begin{figure}
  \centering
  \includegraphics[width=16cm]{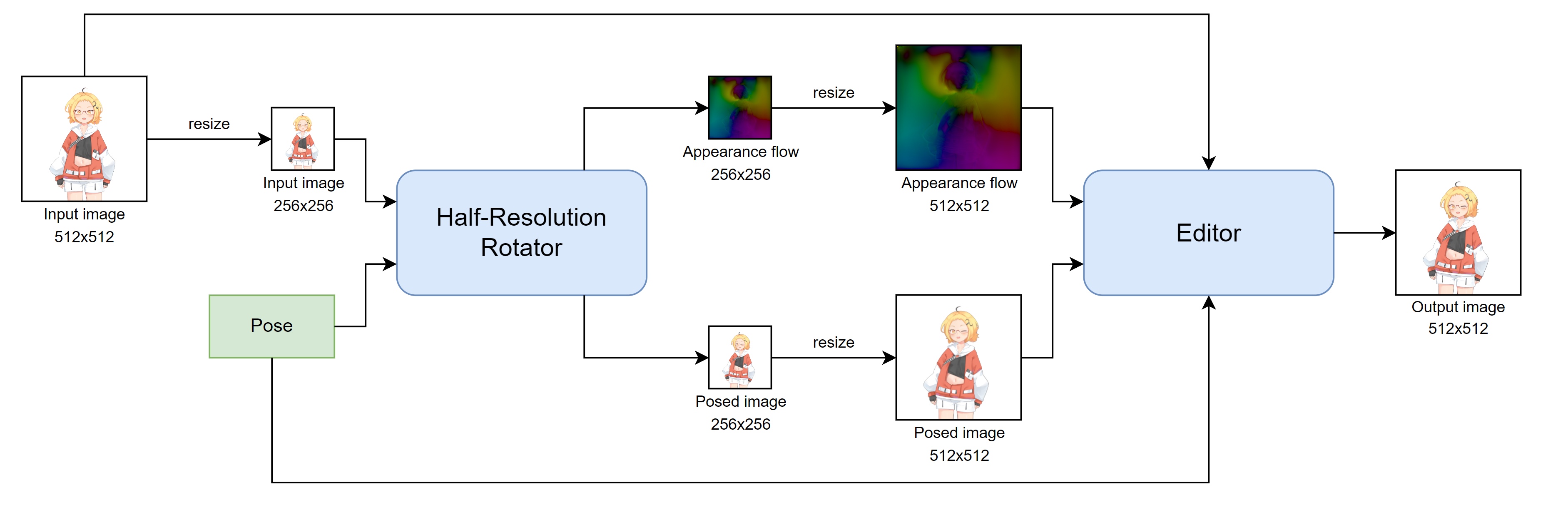}
  \caption{The modified body rotator module.}
  \label{fig:new-body-rotator-architecture}
\end{figure}

There is no change to the interface of the half-resolution rotator. It still takes (1) the input image scaled to $256 \times 256$ and (2) a pose vector, and it still outputs an appearance flow and an image of the posed character, both at the $256 \times 256$ resolution. On the other hand, we slightly change the editor's interface. It now takes in both outputs of the half-resolution rotator, scaled up to $512 \times 512$, instead of taking just the appearance flow like THA3's editor. The overall structure of the modified body rotator is given in in Figure~\ref{fig:new-body-rotator-architecture}.

We changed all backbone networks to U-Nets with attention layers, which are now widely used in diffusion models \cite{Ho:2020,Dhariwal:2021} and prove to be excellent at image generation. We also changed how the networks handle its inputs and outputs.

\begin{itemize}
  \item The half-resolution rotator now generates a half-resolution posed image in three steps. It (1) warps the input image to generate one candidate output, (2) directly generates another candidate output, and (3) alpha blends the two results together. (See the ``image formation'' part of Figure~\ref{fig:student-architecture}.) In THA3, the alpha blending step is missing.
  \item The editor now has to take into account one additional input, so we make it fuse the input image and the two outputs of the half-resolution rotator with a convolution layer before processing the fused tensor with the backbone network. The rest of the network remains the same.
\end{itemize}

We refer the reader to the Appendix~\ref{sec:network-details} for a more complete description of the changes.

\subsection{Training} \label{sec:large-model-training}

We use datasets created from approximately 8,000 controllable 3D anime character models we individually collected from the Internet. Each example in the datasets contains three items: (1) an image of a character in a ``rest'' post, (2) a pose vector, and (3) another image of the same character after being posed according to the pose vector. The training dataset contains 500,000 examples, while the test dataset contains 10,000. The two datasets do not share 3D models, ensuring clean separation between training and test data. Please refer to the write-up of the THA3 project for how to prepare the dataset \cite{Khungurn:2022:THA3}.

We trained the networks using two types of losses: $L_1$ loss and the perceptual content loss \cite{Johnson:2016}. The half-resolution rotator's training is divided into two phases where the first phase only uses the $L_1$ loss, and the second uses both. The editor's training has three phases. In the first two phases, it is trained like a rotator that operates on $512 \times 512$ images. In the last phase, we add to the network units that take into account the outputs of the half-resolution rotator and continue training. Technical details on the training process can be found in Appendix~\ref{sec:training-details}. They include expressions for loss functions, training phases, weight initialization, optimizers, and learning rate schedules.

\subsection{Results}

\subsubsection{Image Quality}

We compare the new body rotator against the old THA3 body rotator. For qualitative comparison, we evaluate the networks by comparing the images they generate against the groundtruth images in the test dataset. We use three metrics for image similarity: (a) peak signal-to-noise ratio (PSNR), (b) structural similarity (SSIM) \cite{Wang:2004}, and (c) LPIPS \cite{Zhang:2018:LPIPS}. The averages of the metrics over the 10,000 examples of the test dataset are reported in Table~\ref{fig:large-models-image-quality-quantitative}. We can see that the use of U-Net with attention and the additional input to the editor improve all the metrics. The LPIPS, in particular, sees an improvement of approximately 30\% over THA3.

\begin{wraptable}{r}{0.6\textwidth}
  \centering
  \begin{tabular}{|l|rrr|}
    \hline
    Network &
    PSNR $(\uparrow)$ &
    SSIM $(\uparrow)$ &
    LPIPS $(\downarrow)$ \\ 
    \hline
    THA3 \cite{Khungurn:2022:THA3} & 22.369330 & 0.909369 & 0.048016 \\
    Section~\ref{sec:new-architecture} & \textcolor{red}{22.962184} & \textcolor{red}{0.919532} & \textcolor{red}{0.033566} \\
    \hline
  \end{tabular}
  \caption{Quantitative comparison of body rotator models' performance.}
  \label{fig:large-models-image-quality-quantitative}
\end{wraptable}

\begin{figure}
  \centering
  \begin{tabular}[t]{c@{\hskip 0.1cm}c@{\hskip 0.1cm}c}
    \footnotesize{Input} &
    \footnotesize{THA3} &
    \footnotesize{This section} \\
    \frame{\adjincludegraphics[width=3cm,trim={{.35\width} {.6\height} {.25\width} {.2\height}},clip]{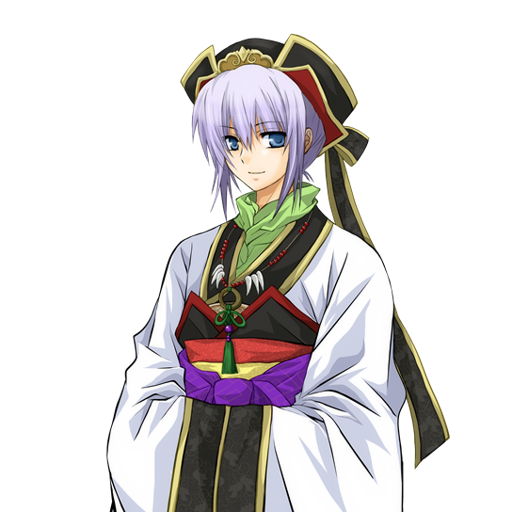}} &
    \frame{\adjincludegraphics[width=3cm,trim={{.35\width} {.6\height} {.25\width} {.2\height}},clip]{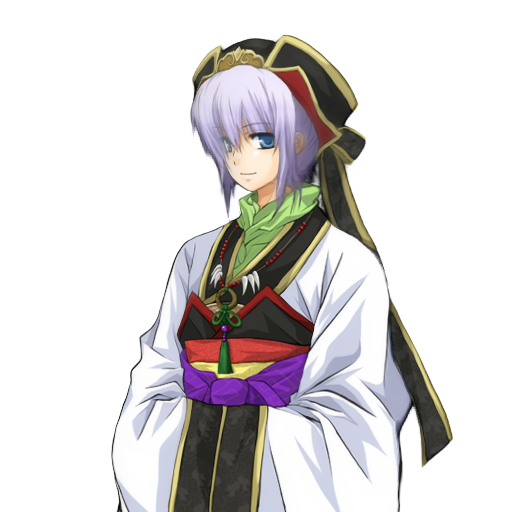}} &
    \frame{\adjincludegraphics[width=3cm,trim={{.35\width} {.6\height} {.25\width} {.2\height}},clip]{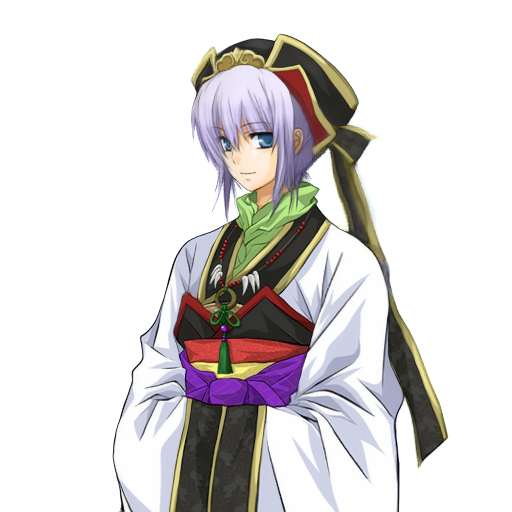}} \\    
    \frame{\adjincludegraphics[width=3cm,trim={{.3\width} {.6\height} {.35\width} {.1\height}},clip]{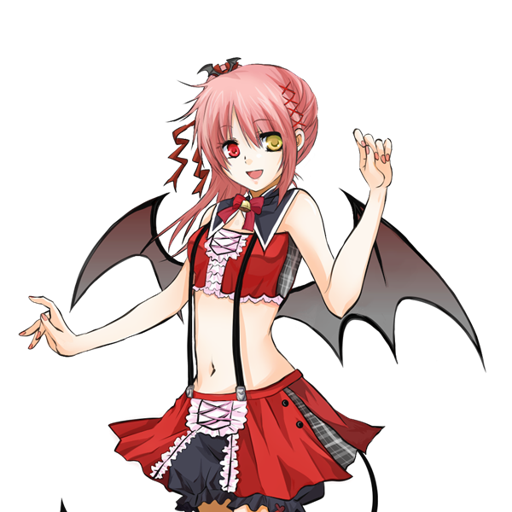}} &
    \frame{\adjincludegraphics[width=3cm,trim={{.3\width} {.6\height} {.35\width} {.1\height}},clip]{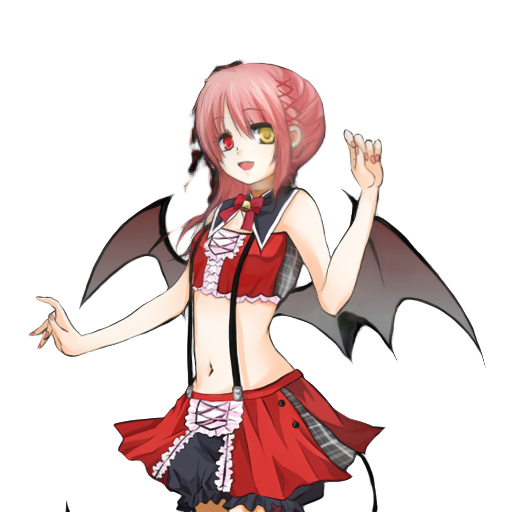}} &
    \frame{\adjincludegraphics[width=3cm,trim={{.3\width} {.6\height} {.35\width} {.1\height}},clip]{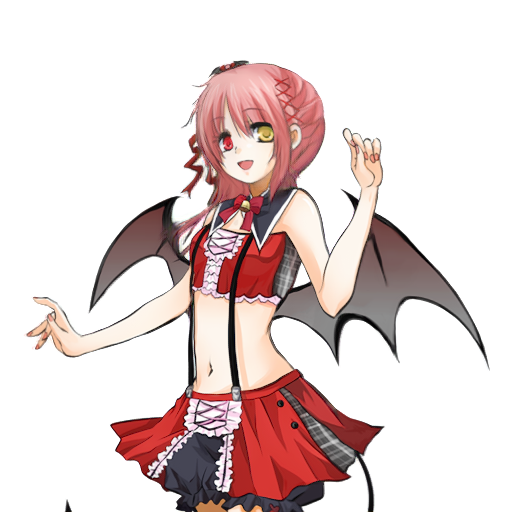}} \\
    \frame{\adjincludegraphics[width=3cm,trim={{.3\width} {.6\height} {.30\width} {.1\height}},clip]{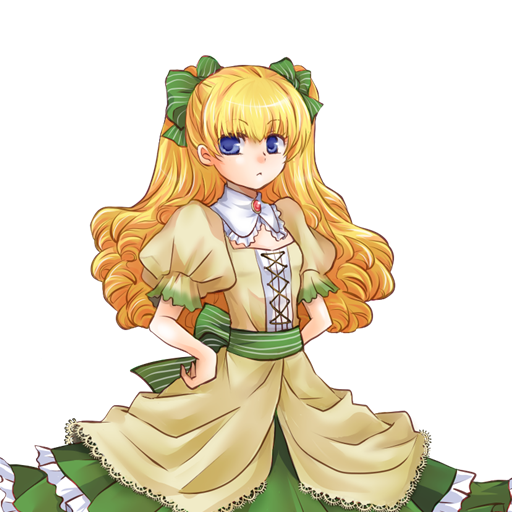}} &
    \frame{\adjincludegraphics[width=3cm,trim={{.3\width} {.6\height} {.30\width} {.1\height}},clip]{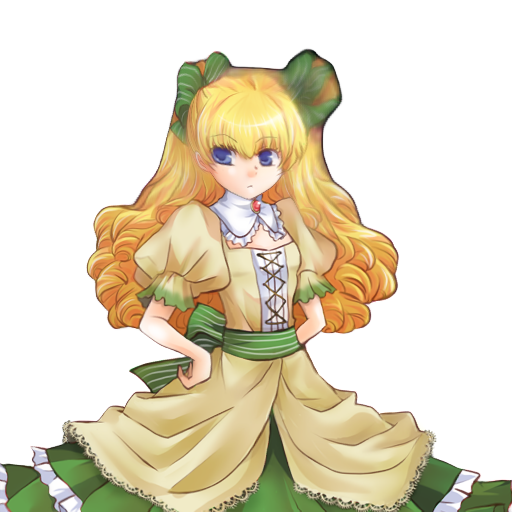}} &
    \frame{\adjincludegraphics[width=3cm,trim={{.3\width} {.6\height} {.30\width} {.1\height}},clip]{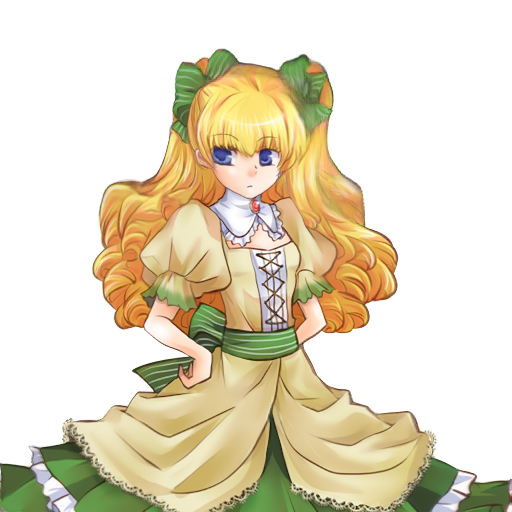}}
  \end{tabular}  
  \caption{Qualitative comparison between images generated by body rotator models. The artworks were created by Mikatsuki Arpeggio \cite{MikatsukiArpeggio, MikatsukiArpeggio:TaoistBoy, MikatsukiArpeggio:KoakumaMei, MikatsukiArpeggio:Marietta}.}
  \label{fig:large-models-image-quality-qualitative}
\end{figure}

For qualitative comparison, we applied the networks to three hand-drawn characters, and we show the results in Figure~\ref{fig:large-models-image-quality-qualitative}. The characters' faces and bodies are rotated to the left of the viewer with the largest possible angles. For the 1\textsuperscript{st} and 2\textsuperscript{nd} characters, we can see that the THA3 rotator could not produce sharp left silhouettes of the faces, and the ribbons worn by the 2\textsuperscript{nd} character are close to being completely erased. On the other hand, the architecture we propose generated much sharper silhouettes and preserved the ribbons better. For the 3\textsuperscript{rd} character, the THA3 rotator generated blurry hair and ribbons on the right side, while ours generated sharper results. We can also see here that our proposed architecture preserved textures in the area better than the baseline.

\subsubsection{Model Size and Speed} \label{sec:large-model-size-and-speed}

Table~\ref{fig:large-models-comparison} compares the size and speed of THA3 system and our proposal. The new editor network is 4 times larger than the THA3 one, but it does not significantly increase the size of the whole system because there are four other networks that are already as large as it is. We assessed the system's speed by measuring the time it takes to fully process one input image and one pose, mirroring the situation in which it is used to generate one animation frame in an application. We performed experiments on three different computers, identified by the letters A to C, whose details are given in Appendix C. The computers have different GPUs, ranging from a research-oriented card to a consumer-oriented gaming one. From Table~\ref{fig:large-models-comparison}, we can see that our proposed architecture, while yielding higher image quality, were about 3 to 4 times slower than the THA3 system. In terms of number of frames per second (FPS), THA3 can in the best case\footnote{FPS inside an application can be lower due to time spent on processing user inputs and updating UI.} achieve around 30 FPS on a machine with very powerful GPUs, but the proposed architecture cannot even make 10 FPS under the same settings.

\begin{table}
  \centering
  \begin{tabular}{|l|r|r|r|r|r|r|}
    \hline
    \multirow{3}{*}{System} & \multicolumn{3}{c|}{Size (MB)} & \multicolumn{3}{c|}{Time needed to generate a frame (ms)} \\ 
    \cline{2-7}
    & \multirow{2}{*}{HRR\footnotemark} & \multirow{2}{*}{Editor} & \multirow{2}{*}{Total\footnotemark} & \multicolumn{1}{c|}{Computer A} & \multicolumn{1}{c|}{Computer B} & \multicolumn{1}{c|}{Computer C} \\ 
    & & & & \multicolumn{1}{c|}{\footnotesize{(RTX A6000)}} & \multicolumn{1}{c|}{\footnotesize{(Titan RTX)}} & \multicolumn{1}{c|}{\footnotesize{(GTX 1080 Ti)}} \\ 
    \hline 
    THA3 \cite{Khungurn:2022:THA3} & 128 & 33 & 517 & 35.899 & 41.409 & 64.607 \\
    This section & 136 & 137 & 627   & 125.843 & 116.763 & 159.647 \\
    \hline
  \end{tabular}
  \caption{Size and speed comparison between the THA3 system and our proposed one. The times needed to generate a frame are averages of 1,000 measurements.}
  \label{fig:large-models-comparison}
\end{table}

\section{Distillation} \label{sec:distillation}

Results from the last section reveal that our quest to improve image quality results in a much slower model. Our next task, then, is to improve image generation speed so that real-time performance is achieved on less powerful hardware.

We first observe that the system is overly capable. At any time, we can change the input image, and the change would be reflected on the output image immediately. Nevertheless, in computer games or in streaming, a character does not change its appearance every second or every minute. This functionality is thus unnecessary. By creating a model specialized to a particular input image, we may obtain a faster model that works under real-time constraints. If we prepare many such models in advance, we may swap the models to change characters or allow a character to change its appearance when needed.

To create such a specialized model, we rely on {\bf knowledge distillation} \cite{Hinton:2015}, which is the practice of training a smaller model (the {\bf student}) to mimic the behavior of a larger model (the {\bf teacher}). In our case, the teacher is the full system as proposed in the last section.

All of the student models we proposed are coordinate-based networks \cite{Tancik:2020} because, by construction, they allow generating any specific subimage at a cost proportional to the subimage's size. Moreover, unlike CNN-based image generators, subimage generation can be done without having to generate the whole image. This feature is beneficial for game characters and real-time streaming because, in some cases, the user might want to depict only the head instead of the whole torso. The specific architecture we employ is the SInusoidal REpresentation Network (SIREN) \cite{Sitzmann:2020} because we found that it produced smooth images that fit well with the anime style. On the other hand, a competing approach (ReLU MLP with Fourier features \cite{Tancik:2020}) tend to produce grainy artifacts \cite{Sitzmann:2020}.

\footnotetext[4]{HRR stands for ``half-resolution rotator.''}
\footnotetext[5]{Complete THA systems have three other networks. This column contains the sizes of all the networks combined.}

\subsection{Student Architecture} \label{sec:student-architecture}

\begin{figure}
  \centering
  \includegraphics[width=12cm]{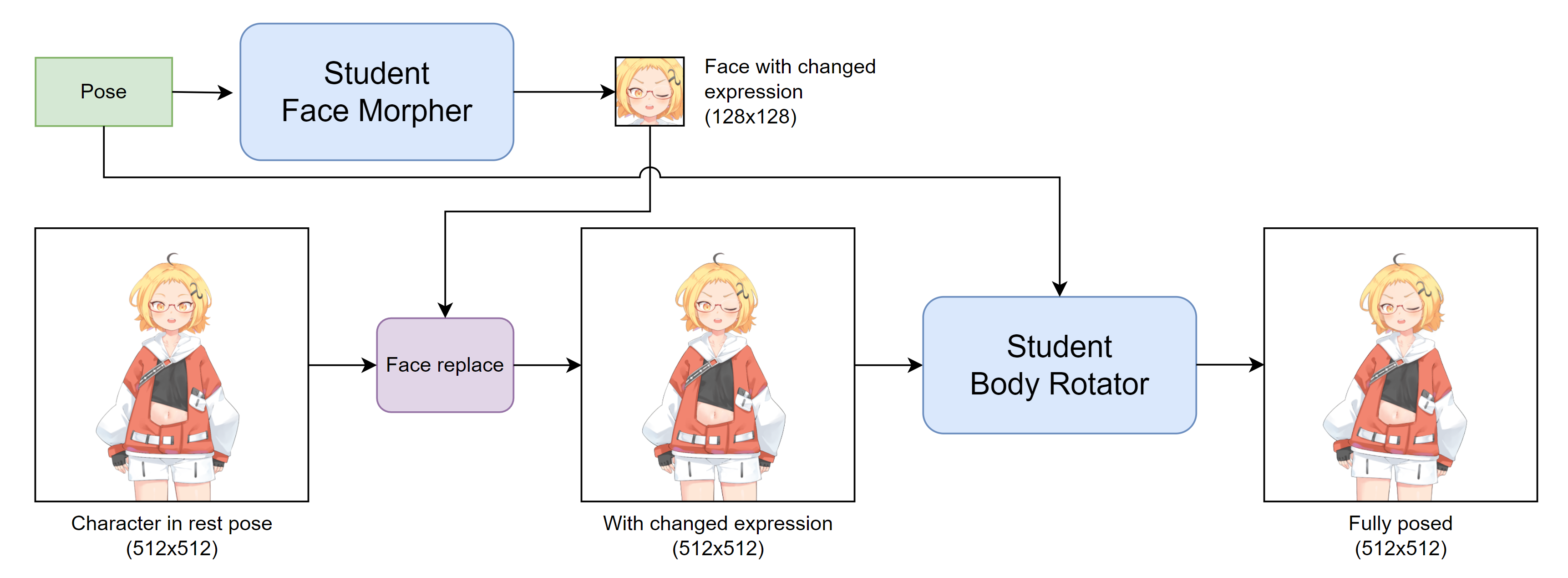}
  \caption{Overall architecture of the student model.}
  \label{fig:student-architecture}
\end{figure}

Like the full system, the student model contains two modules, {\bf the face morpher} and the {\bf body rotator}, with the same functionality. Instead of being a collection of five big networks, they are now two small networks whose total size is less than 2 MB. An overview of the student's architecture is shown in Figure~\ref{fig:student-architecture}.

\begin{figure}
  \centering
  \includegraphics[width=12cm]{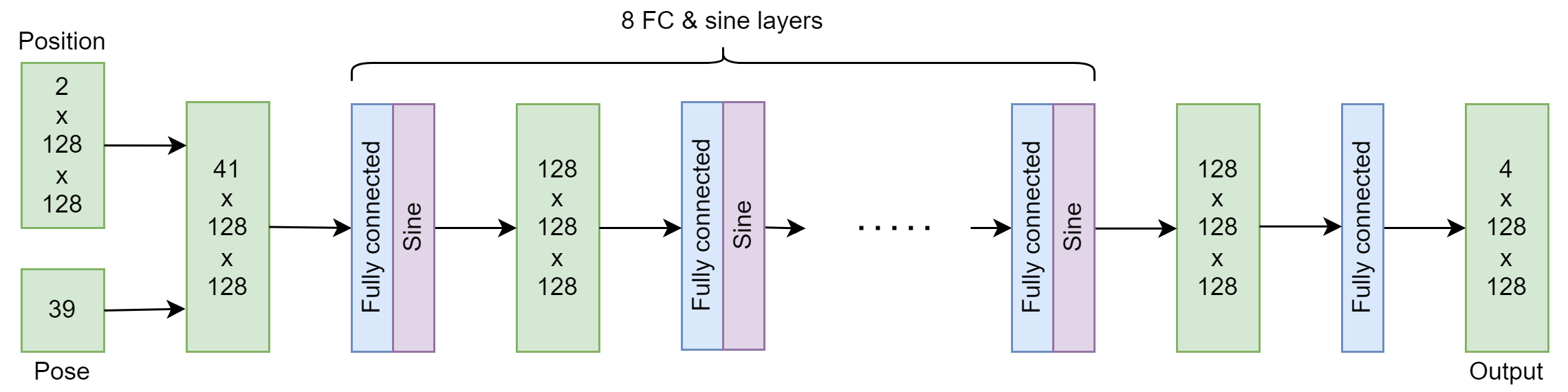}
  \caption{Architecture of the student face morpher.}
  \label{fig:student-face-morpher-architecture}
\end{figure}

The student face morpher is a SIREN with 9 fully connected layers where each hidden layer has 128 neurons. Its architecture is depicted in Figure~\ref{fig:student-face-morpher-architecture}. It is trained to generate a $128 \times 128$ area of the input image that contains the character's movable facial organs (eyebrows, eyes, mouth, and jaw). The SIREN receives as input a pixel position (2 dimensions) and a facial pose (39 dimensions), and it produces an RGBA pixel (4 dimensions). Its size is only 475 KB.

\begin{figure}
  \centering
  \includegraphics[width=12cm]{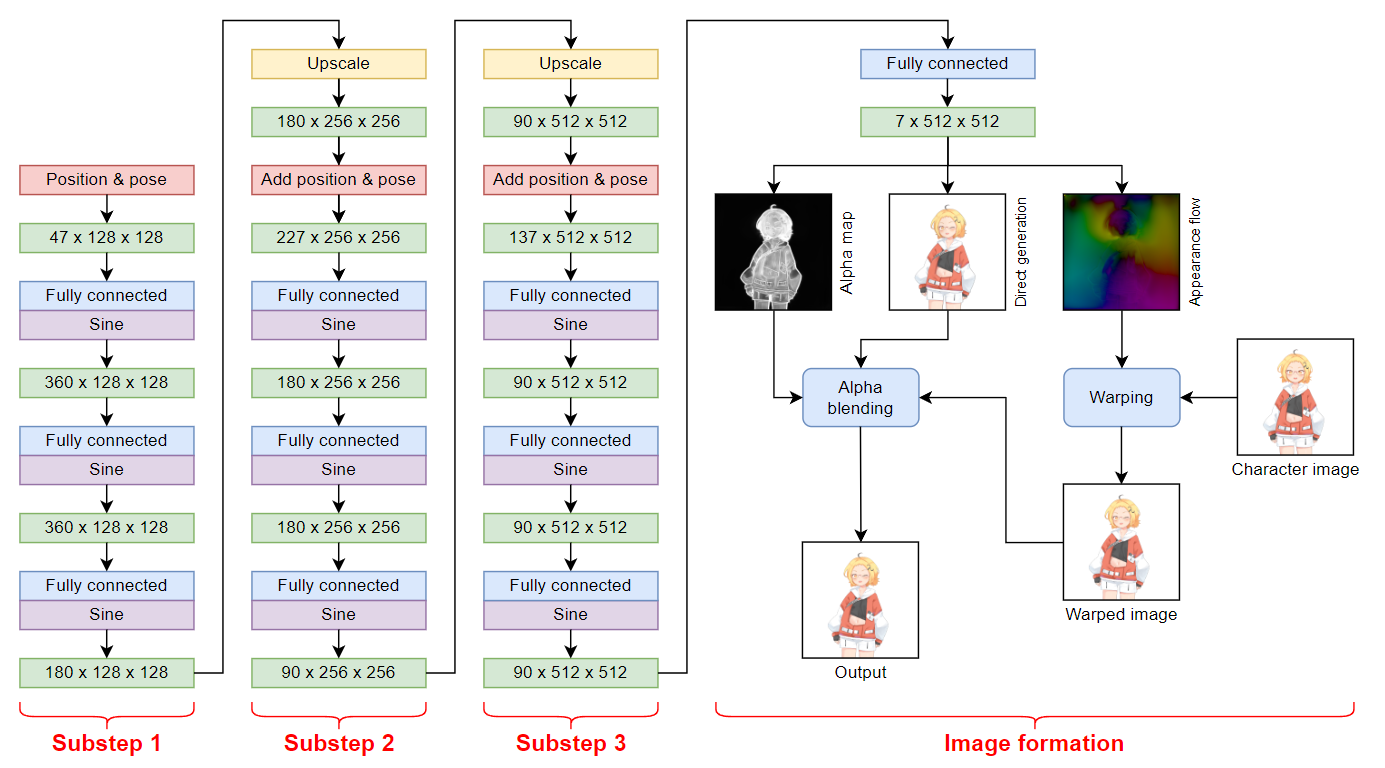}
  \caption{Architecture of the student body rotator.}
  \label{fig:student-body-rotator-architecture}
\end{figure}

The student body rotator's architecture is more complicated because it needs to generate much larger outputs ($512 \times 512$ images) in real time. A vanilla SIREN like the student face morpher would be too slow because it has to operate on tensors with a spatial size of $512 \times 512$ at all of its layers. To improve speed, we divide the image generation process into three substeps where the network would operate on tensors with spatial resolutions of $128 \times 128$ first, then $256 \times 256$, and lastly $512 \times 512$. Each substep has 3 fully connected layers, except for the last one which has 4, resulting in a network with 10 such layers. Moreover, in order to preserve fine details of the input image, the network does not generate the output image directly. Instead, it uses the image formation process employed by the teacher's body rotator. In particular, the network is trained to generate (1) an appearance flow, (2) an RGBA image, and (3) an alpha map. The output image is formed by first using the appearance flow to warp the input character image and then alpha blending the warped image with the directly generated RGBA image. The model size's is about 1.3 MB. The architecture of the student body rotator is depicted in Figure~\ref{fig:student-body-rotator-architecture}.

\subsection{Student Training}

\subsubsection{Face Morpher}

The student face morpher is trained to minimize the L1 differences between its outputs and those generated by the teacher face morpher. The loss function has two terms. The first is the L1 difference between the whole outputs, and the second is the L1 difference between areas that contain movable facial parts. We weigh the second term 20 times more than the first because the movable parts are small compared to the whole face. The precise definition of the loss is given in Appendix~\ref{sec:student-training-details}.

At training time, the character image is fixed, and the pose vectors are sampled uniformly from the training dataset of the full system. Training lasts 2 epochs (1M examples), and the batch size is 8. We use the Adam optimizer with $\beta_1 = 0.9$ and $\beta_2 = 0.999$. The learning rate starts at $10^{-4}$ and decays to $3.33 \times 10^{-5}$, $1 \times 10^{-5}$ and then $3.33 \times 10^{-6}$ after 200K, 500K, and 800K training examples, respectively. Training takes about an hour and a half on a computer with four V100 GPUs.

\subsubsection{Body Rotator}

Recall first that the body rotator uses the same image formation process as the teacher body rotator. The outputs of the last fully-connected layer are (1) an appearance flow $I_{\mrm{flow}}$ which is immediately used to generate a warped image $I_{\mrm{warped}}$ from the fixed character image, (2) an RGBA image $I_{\mrm{direct}}$, and (3) an alpha map $I_{\mrm{alpha}}$. Then, $I_{\mrm{warped}}$, $I_{\mrm{direct}}$ and $I_{\mrm{alpha}}$ are then combined with alpha blending to generate the final output image $I_{\mrm{final}}$. Because the teacher also generates these data as well, we distinguish between those generated by the student with the superscript ``$S$'' (such as $I_{\mrm{flow}}^S$, $I_{\mrm{warped}}^S$) and those generated by the teacher with the superscript ``$T$'' (such as $I_{\mrm{direct}}^T$, $I_{\mrm{final}}^T$).

The student body rotator is trained to minimize a 4-termed loss where each term involves one of the generated data above:
\begin{align*}
  \mcal{L}_{\mrm{br}} = \lambda_{\mrm{flow}} \mcal{L}_{\mrm{flow}} + \lambda_{\mrm{warped}} \mcal{L}_{\mrm{warped}} + \lambda_{\mrm{direct}} \mcal{L}_{\mrm{direct}} + \lambda_{\mrm{final}} \mcal{L}_{\mrm{final}} \label{eqn:student-body-rotator-loss}
\end{align*}
where $\mcal{L}_{\square} = \| I_{\square}^S - I_{\square}^T \|_1$, and $\square$ can be replaced with the suffixes in the above equation. The $\lambda$-variables are weights that change throughout the training process, which is divided into three phases as shown in Table~\ref{fig:student-body-rotator-training-phases}. We can see that the the first phase focuses on training the direct generation, the second on the warping, and the third on the final output.

\begin{wraptable}{*}{0.55\textwidth}
  \centering
  \begin{tabular}{|c|c|r|r|r|r|}
    \hline
    Phase & \# Examples & $\lambda_{\mrm{flow}}$ & $\lambda_{\mrm{warped}}$ & $\lambda_{\mrm{direct}}$ &  $\lambda_{\mrm{final}}$ \\
    \hline
    \#1 & $\leq$ 400K & 0.50 & 0.25 & 2.00 & 0.25 \\
    \#2 & $\leq$ 800K & 5.00 & 2.50 & 1.00 & 1.00 \\
    \#3 & $\leq$ 1.5M & 1.00 & 1.00 & 1.00 & 10.00 \\
    \hline
  \end{tabular}
  \caption{Training phases of the student body rotator.}
  \label{fig:student-body-rotator-training-phases}
\end{wraptable}

Much like what we do with the student face morpher, we also sample pose vectors from the training dataset of the full system, use the Adam optimizer with $\beta_1 = 0.9$ and $\beta_2 = 0.999$, and set the batch size to 8. However, training now lasts for 3 epochs (1.5M examples). Learning rate starts from $10^{-4}$ and decays to $3 \times 10^{-5}$, $10^{-5}$, and $3 \times 10^{-6}$ after we have shown 200K, 600K, and 1.3M training examples, respectively. Training takes about 10 hours on a computer with four V100 GPUs. We have not measured training time on other computers, but we surmise that it would take several ten hours on a machine with a single GPU.

\subsection{Results}

We assess a model by using it to pose characters according to 1,000 fixed poses taken from the test dataset in Section~\ref{sec:large-model-training}. For each posed image, we compute the PSNR with respect to the corresponding image generated by the teacher model. We then record the average of the 1,000 resulting PSNR values.

\subsubsection{Comparison Against the Teacher}

\begin{table}
  \centering
  \begin{tabular}{|l|r|r|r|}
    \hline
    \multirow{3}{*}{System} & \multicolumn{3}{c|}{Time needed to generate a frame (ms)} \\ 
    \cline{2-4}
    & \multicolumn{1}{c|}{Computer A} & \multicolumn{1}{c|}{Computer B} & \multicolumn{1}{c|}{Computer C} \\ 
    & \multicolumn{1}{c|}{\footnotesize{(RTX A6000)}} & \multicolumn{1}{c|}{\footnotesize{(Tital RTX)}} & \multicolumn{1}{c|}{\footnotesize{(GTX 1080 Ti)}} \\ 
    \hline 
    THA3 \cite{Khungurn:2022:THA3} & 35.899 & 41.409 & 64.607 \\
    Section~\ref{sec:new-architecture} & 125.840 & 116.760 & 159.640\\
    Student model & \textcolor{red}{12.523} & \textcolor{red}{15.098} & \textcolor{red}{22.091} \\
    \hline
  \end{tabular}
  \caption{Comparison between average time required to generate a frame of animation by the THA3 system, the teacher model (Section~\ref{sec:new-architecture}), and the student model.}
  \label{fig:student-teacher-running-time-comparison}
\end{table}

\begin{wraptable}{r}{0.4\textwidth}
  \centering
  \begin{tabular}{|l|r|}
    \hline
    Character & PNSR (dB) \\
    \hline 
    Top \cite{MikatsukiArpeggio:TaoistBoy}  & 36.156 \\
    Middle \cite{MikatsukiArpeggio:KoakumaMei} & 36.048 \\
    Bottom \cite{MikatsukiArpeggio:Marietta} & 34.543 \\ 
    \hline  
  \end{tabular}
  \caption{Average PSNR of images generated by student models trained to animate the characters from Figure~\ref{fig:large-models-image-quality-qualitative}.}
  \label{fig:student-teacher-image-quality}
\end{wraptable}

As mentioned in Section~\ref{sec:student-architecture}, the student model is much smaller than the full system: 1.8 MB versus 627 MB. It is also around 8 times faster to generate a single animated frame as can be seen in Table~\ref{fig:student-teacher-running-time-comparison}. Compared to the THA3 system, it is about 3 times faster and thus can now achieve real-time animation ($\geq$ 30 FPS) on Computer C, which has a standard consumer GPU.

As for the quality of generated images, we trained student models on the three characters in Figure~\ref{fig:large-models-image-quality-qualitative}. We report the averaged PSNRs in Table~\ref{fig:student-teacher-image-quality}, which range from 34 dB to around 36 dB. This means that the average error is about 2\% of the maximum pixel value. Qualitatively, it is hard to spot large differences between outputs of the students and the teacher, but a student model might ignore extremely fine details such as the black dot that represents the nose as can be seen in Figure~\ref{fig:student-architecture-ablation-qualitative} and Figure~\ref{fig:student-training-ablation-qualiative-00}.

\subsubsection{Ablation Study on Student Network Architecture}

\begin{figure}
  \centering
  \begin{tabular}[t]{c@{\hskip 0.1cm}c@{\hskip 0.1cm}c@{\hskip 0.1cm}c}
    \footnotesize{Vanilla SIREN} &
    \footnotesize{Section~\ref{sec:student-architecture} w/o multi-res} &
    \footnotesize{Section~\ref{sec:student-architecture}} &
    \footnotesize{Teacher (Section \ref{sec:new-architecture})} \\
    \frame{\adjincludegraphics[width=3.9cm,trim={{.15\width} {.0\height} {.15\width} {.1\height}},clip]{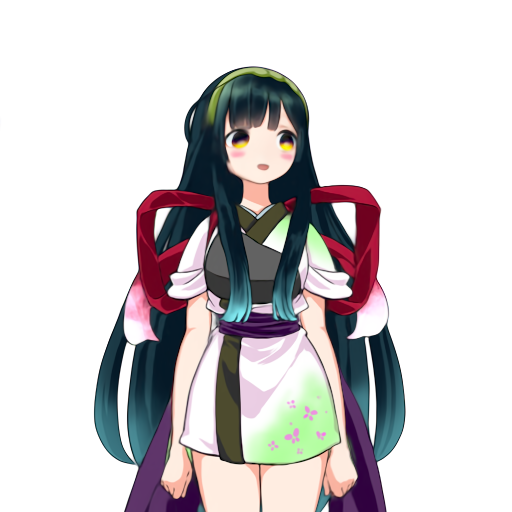}} &
    \frame{\adjincludegraphics[width=3.9cm,trim={{.15\width} {.0\height} {.15\width} {.1\height}},clip]{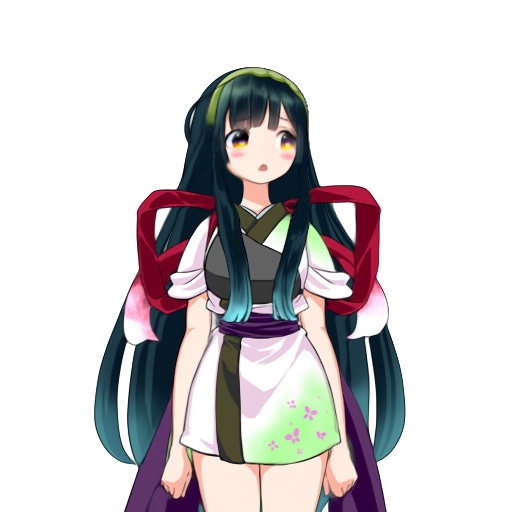}} &
    \frame{\adjincludegraphics[width=3.9cm,trim={{.15\width} {.0\height} {.15\width} {.1\height}},clip]{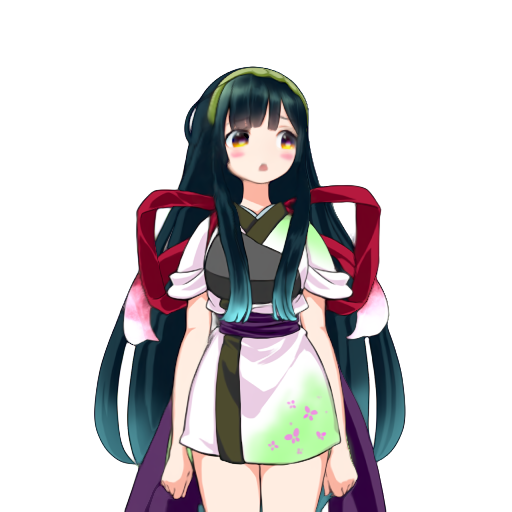}} &
    \frame{\adjincludegraphics[width=3.9cm,trim={{.15\width} {.0\height} {.15\width} {.1\height}},clip]{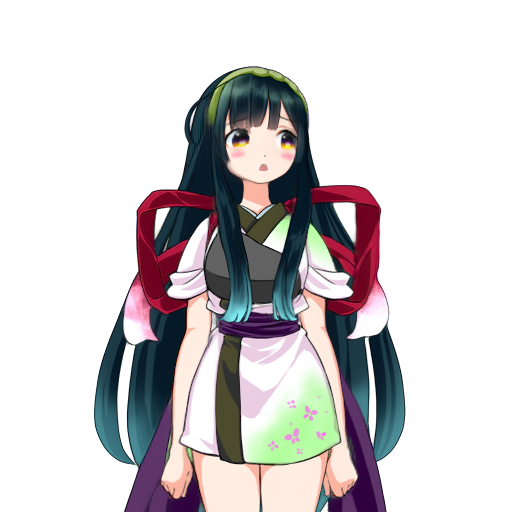}} \\
    \frame{\adjincludegraphics[width=3.9cm,trim={{.43\width} {.65\height} {.41\width} {.21\height}},clip]{images/student-architecture-ablation/mode_16.png}} &
    \frame{\adjincludegraphics[width=3.9cm,trim={{.43\width} {.65\height} {.41\width} {.21\height}},clip]{images/student-architecture-ablation/mode_17.png}} &
    \frame{\adjincludegraphics[width=3.9cm,trim={{.43\width} {.65\height} {.41\width} {.21\height}},clip]{images/student-architecture-ablation/mode_14.png}} &
    \frame{\adjincludegraphics[width=3.9cm,trim={{.43\width} {.65\height} {.41\width} {.21\height}},clip]{images/student-architecture-ablation/mode_07.png}}    
  \end{tabular}  
  \caption{Qualitative comparison between images generated by the teacher and three student architectures. The character is \copyright~Touhoku Zunko $\cdot$ Zundamon Project \cite{ZunZunProject}.}
  \label{fig:student-architecture-ablation-qualitative}
\end{figure}

In this section, we show that the proposed architecture yielded improvement over simpler alternatives. We compare our architecture against (a) a vanilla SIREN that generates the output image directly, and (b) our proposed architecture where the body rotator is modified so that it always operates at the $512 \times 512$ resolution. We trained the three architectures to animate a specific character image \cite{TouhokuZunko00}, and we evaluated them with the average PSNR metric. We also measured the average time it took to generate an animation frame on the 3 computers used in Section~\ref{sec:large-model-size-and-speed}.  The statistics are given in Table~\ref{fig:student-architecture-ablation-quantitative}. For qualitative comparison, we show generated images in Figure~\ref{fig:student-architecture-ablation-qualitative}.

\begin{table}
  \centering
  \begin{tabular}{|l|r|r|r|r|}
    \hline
    \multirow{3}{*}{Architecture} & \multirow{3}{*}{PSNR (dB)} & \multicolumn{3}{c|}{Time needed to generate a frame (ms)} \\     
    \cline{3-5}
    & & \multicolumn{1}{c|}{Computer A} & \multicolumn{1}{c|}{Computer B} & \multicolumn{1}{c|}{Computer C} \\ 
    & & \multicolumn{1}{c|}{\footnotesize{(RTX A6000)}} & \multicolumn{1}{c|}{\footnotesize{(Tital RTX)}} & \multicolumn{1}{c|}{\footnotesize{(GTX 1080 Ti)}} \\ 
    \hline 
    \footnotesize{Vanilla SIREN} & 38.259 & 21.319 & 33.086 & 54.937 \\
    \footnotesize{Section~\ref{sec:student-architecture} w/o multi-res} & \textcolor{red}{38.923} & 24.337 & 34.883 & 57.394 \\
    \footnotesize{Section~\ref{sec:student-architecture}} & 38.881 & \textcolor{red}{12.523} & \textcolor{red}{15.098} & \textcolor{red}{22.091} \\
    \hline
  \end{tabular}
  \caption{An ablation study on the architecture of the student model.}
  \label{fig:student-architecture-ablation-quantitative}
\end{table}

We see in Table~\ref{fig:student-architecture-ablation-quantitative} that the student architectures' PSNR values are comparable to one another. However, Figure~\ref{fig:student-architecture-ablation-qualitative} reveals that the vanilla SIREN architecture is qualitatively much worse than the other two because it cannot reproduce fine face details, such as the eyebrows, the mouth shapes, and the highlights on the pupils. Preserving these fine details necessitates the more complicated image formation steps. The architecture without multi-resolution SIREN is slightly more accurate than the proposed architecture. However, it is very hard to spot differences between their generated images in Figure~\ref{fig:student-architecture-ablation-qualitative}. The advantage of the proposed architecture is its speed: Table~\ref{fig:student-architecture-ablation-quantitative} shows that it is two times faster than the architecture without multi-resolution SIREN. In other words, the multi-resolution design maintains accuracy while making the network significantly faster.

\subsubsection{Ablation Study on Student Training Process}

\begin{wraptable}{r}{0.4\textwidth}
  \centering
  \begin{tabular}{|l|c|c|c|r|}
    \hline
    \multirow{2}{*}{Model} & \multicolumn{3}{c|}{Training phases} & \multirow{2}{*}{PSNR (dB)} \\
    \cline{2-4}
    & \#1 & \#2 & \#3 & \\
    \hline
    A & & & \checkmark & 29.308 \\
    B & & \checkmark & & 29.118 \\
    C & & \checkmark & \checkmark & 29.484 \\
    D & \checkmark & & & 38.026 \\
    E & \checkmark & & \checkmark & 38.668 \\
    F & \checkmark & \checkmark & & 37.399 \\
    G & \checkmark & \checkmark & \checkmark & \textcolor{red}{38.881} \\
    \hline
  \end{tabular}
  \caption{Quantitative comparison between student models trained with and without specific training phases.}
  \label{fig:student-training-ablation-quantitative}
\end{wraptable}

\begin{figure}
  \centering
  \begin{tabular}{c@{\hskip 0.1cm}c@{\hskip 0.1cm}c@{\hskip 0.1cm}c}
    Model A & Model B & Model C & Model G (Sec~\ref{sec:student-architecture}) \\
    \frame{\adjincludegraphics[width=3.9cm,trim={{.3\width} {.6\height} {.3\width} {.1\height}},clip]{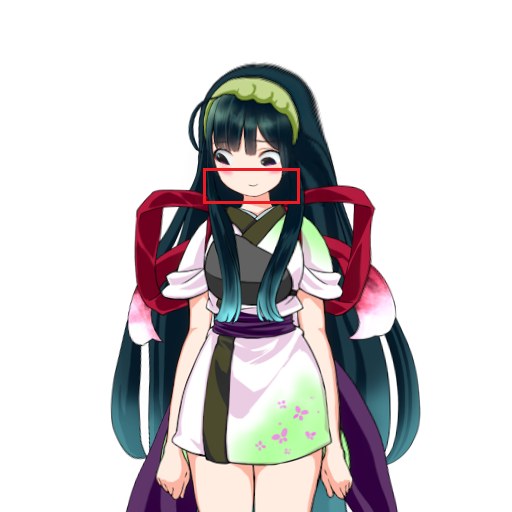}} &
    \frame{\adjincludegraphics[width=3.9cm,trim={{.3\width} {.6\height} {.3\width} {.1\height}},clip]{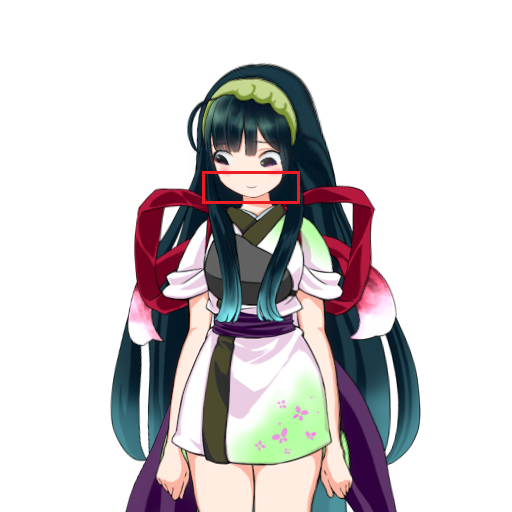}} &
    \frame{\adjincludegraphics[width=3.9cm,trim={{.3\width} {.6\height} {.3\width} {.1\height}},clip]{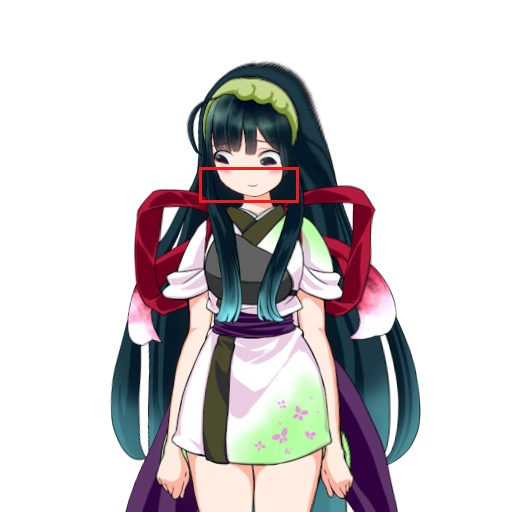}} &
    \frame{\adjincludegraphics[width=3.9cm,trim={{.3\width} {.6\height} {.3\width} {.1\height}},clip]{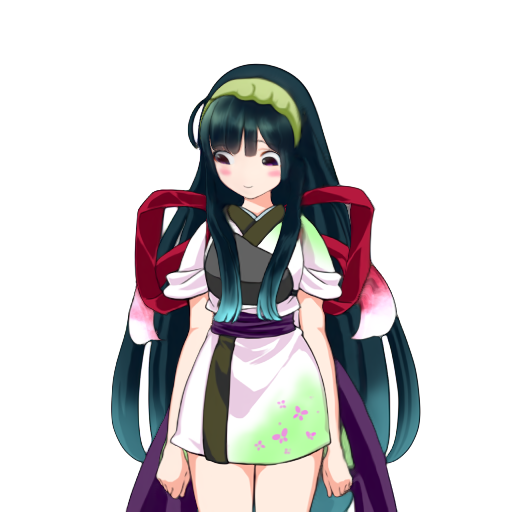}} \\
    Model D & Model E & Model F & Teacher \\
    \frame{\adjincludegraphics[width=3.9cm,trim={{.3\width} {.6\height} {.3\width} {.1\height}},clip]{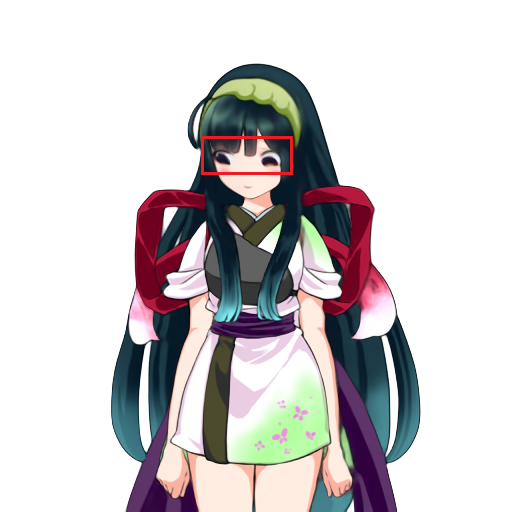}} &
    \frame{\adjincludegraphics[width=3.9cm,trim={{.3\width} {.6\height} {.3\width} {.1\height}},clip]{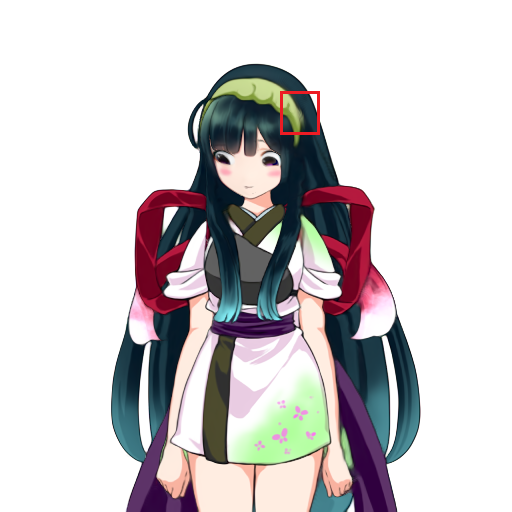}} &
    \frame{\adjincludegraphics[width=3.9cm,trim={{.3\width} {.6\height} {.3\width} {.1\height}},clip]{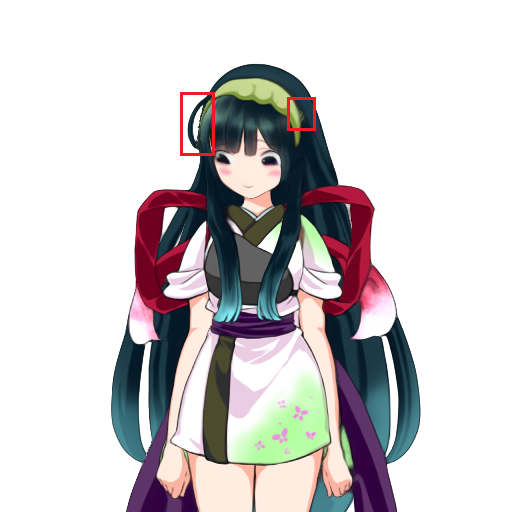}} &
    \frame{\adjincludegraphics[width=3.9cm,trim={{.3\width} {.6\height} {.3\width} {.1\height}},clip]{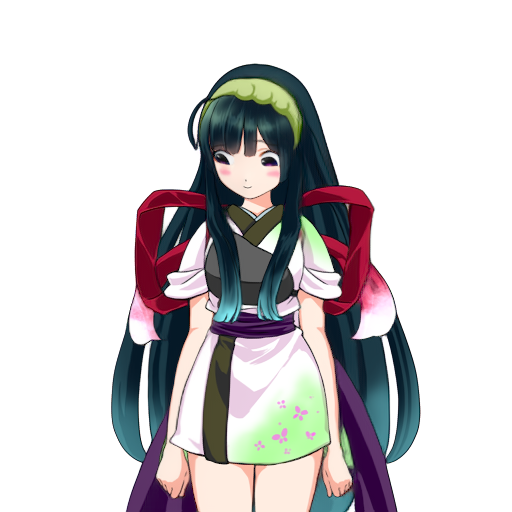}}    
  \end{tabular}
  \caption{Qualitative comparison between outputs of models in Table~\ref{fig:student-training-ablation-quantitative}. Problematic areas are highlight with red rectangles.}
  \label{fig:student-training-ablation-qualiative-00}
\end{figure}

The training process for the student model has 3 training phases with different weights on loss terms. To show the necessity of the phases, we trained student models on the character image in the last section, ablating the training phases while keeping the rest of the settings the same. We report the models' average PSNR values in Table~\ref{fig:student-training-ablation-quantitative}. We see that employing all phases yielded the best score. Omitting Phase \#1 resulted in significantly worse image quality. This manifests qualitatively as noticeable differences in the shape of the rotated faces in Figure~\ref{fig:student-training-ablation-qualiative-00}. Models that were trained with Phase \#1 have PSNR scores of around 38, showing that they approximate the teacher's overall outputs well. However, there are visible degradations in the details. Model D did not reproduce the highlights on the pupils. Model E and Model F produced artifacts around the headband. Moreover, Model F also yielded jagged edges on one side of the head. Model G, which experienced all training phases, achieved the highest PSNR score and produced the least amount of artifacts, showing the necessity of all the training phases.

\subsubsection{Miscellaneous Results}

Student models are fast and lightweight that they can be executed inside a web browser and still generate animation in real time. In the supplementary material, we include two demo web applications. One allows the user to pose characters by manipulating UI widgets. The other makes characters imitate the user's movement as captured by a web camera.

\section{Conclusion}

We proposed improvements to the Talking Head Anime 3 (THA3) system, increasing its image quality and speeding it up so that it can generate smooth animation in real time with a consumer gaming GPU. The latter improvement makes the system practical as a streaming tool for the first time. 
The main insight is that we can use a more expensive architecture (U-Net with attention) to get better image quality and then distill the improved model to small and fast students. Our technical contribution includes an effective architecture for the student model (multi-resolution SIREN with warping and blending) and an algorithm to train it. 

There are still several limitations to our work. The image quality, while greatly improved by the new architecture for the body rotator, can still be improved further. The system can only move facial organs  rotate of the face and the torso by small angles. Lastly, while the student model is lightweight enough to run on a consumer gaming GPU, it can still cannot be run on devices such as tablets or mobile phones. We hope to address these problems in future works.

\bibliographystyle{acm}
\bibliography{paper}  

\appendix

\section{Full System's Architecture Details} \label{sec:network-details}

\subsection{U-Net with Attention}

The new backbones of the half-resolution rotator and editor are U-Net with attention \cite{Ho:2020}, which are frequently used in diffusion models. We base our architecture on conditional U-Nets in the diffusion autoencoder paper by Preechakul \etal~\cite{Preechakul:2022}. There, the U-Net takes as input a feature tensor, a time value, and a 1D conditioning vector. The time value and the conditioning vector are mingled tensors derived from the input feature tensor through adaptive instance normalization (AdaIN) units \cite{Huang:2017:AdaIN} that are parts of residual blocks \cite{He:2016:ResNet}. A residual block would have two AdaIN units that are applied in succession: the first for time and the second for conditioning vector. In the diffusion autoencoder paper, the conditioning vector is a $512$-dimensional vector. In our case, the conditioning vector is the 6-dimensional pose vector.\footnote{While the full pose vector has 45 parameters, only 6 that concern the movement of the body are relevant to the networks that we modify.} For our networks, the time value is always $0$ and is totally redundant. We kept the code related to time embedding in place in order to reduce implementation effort. 

\begin{table}
  \centering
  \begin{tabular}{|l|l|l|}    
    \hline
    {\bf Hyperparameter} & {\bf Half-resolution rotator} & {\bf Editor} \\
    \hline
    image resolution & $256 \times 256$ & $512 \times 512$ \\
    \# base channels & 64 & 32 \\
    channel multipliers & 1, 2, 4, 4, 4 & 1, 2, 4, 8, 8, 8 \\    
    \# residual blocks per level & 1 & 1 \\
    \# bottleneck residual blocks & 4 & 4 \\
    resolution with attention blocks & 16 & 16 \\
    \# attention heads & 8 & 8 \\
    \hline
  \end{tabular}
  \caption{Configurations of the U-Net with attention backbones for the half-resolution rotator and editor.}
  \label{fig:unet-configs}
\end{table}

The configurations for the backbone networks are given in Table~\ref{fig:unet-configs}. Both networks scale down the feature tensors to $16 \times 16$ before scaling them back up to the original resolution. Attention blocks are only present at the $16\times16$ resolution. The bottleneck part of each network has 4 residual blocks alternating with three attention blocks. (In other words, there are $3+2 = 5$ attention blocks in total.) Each attention block has 8 attention heads.

The half-resolution rotator and the editor differ not only on the configurations of their backbones but how the backbones are ``wrapped'' by extra units to that they conform to the networks' interfaces. We discuss these extra units in the two following subsections.

\subsection{Half-Resolution Rotator} \label{sec:half-resolution-rotator-architecture}

\begin{figure}
  \centering
  \includegraphics[width=16cm]{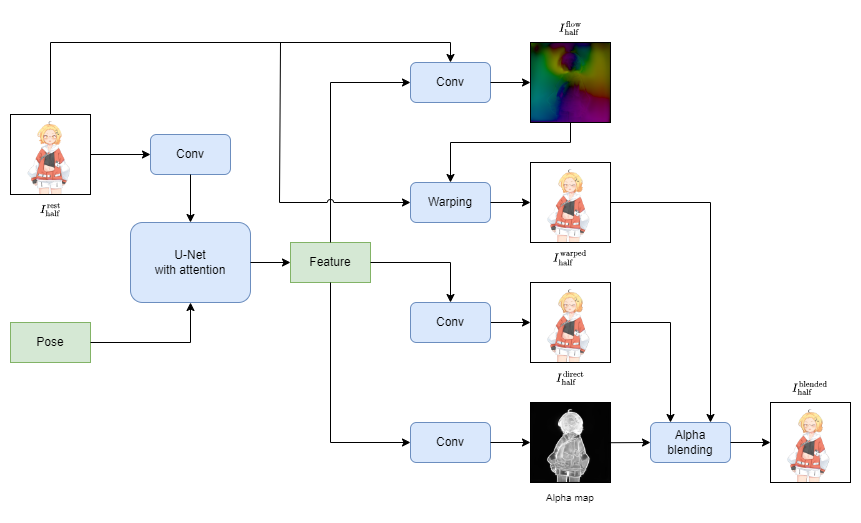}
  \caption{The new half-resolution rotator.}
  \label{fig:half-resolution-rotator-architecture}
\end{figure}

The half-resolution rotator is depicted in Figure~\ref{fig:half-resolution-rotator-architecture}. It takes as input
\begin{enumerate}
  \item $I^{\mrm{rest}}_{\mrm{half}}$, a $4 \times 256 \times 256$ RGBA image of the character in rest pose obtained by downscaling the original input image, and
  \item $\mathbf{p}$, a 6-dimensional pose vector.
\end{enumerate}
Because the U-Net with attention backbone takes in a $64 \times 256 \times 256$ tensor as input, $I^{\mrm{rest}}_{\mrm{half}}$ must be converted to this shape with a convolution layer. The backbone produces another $64 \times 256 \times 256$ feature tensor, which is then used to perform several image processing operations. (See Section~\ref{sec:baseline}.)
\begin{itemize}
  \item \emph{Warping.} The feature tensor is passed to a convolution layer to produce an appearance flow $I^{\mrm{flow}}_{\mrm{half}}$ of size $2 \times 256 \times 256$. It is then used to warp the input image ($I^{\mrm{rest}}_{\mrm{half}}$) to produce a warped image $I^{\mrm{warped}}_{\mrm{half}}$.
  
  \item \emph{Direct generation.} The feature tensor is converted to a $4 \times 256 \times 256$ RGBA image, denoted by $I^{\mrm{direct}}_{\mrm{half}}$.

  \item \emph{Blending.} The feature tensor is converted to a $1 \times 256 \times 256$ alpha map, which is then used to blend the warped image $I^{\mrm{warped}}_{\mrm{half}}$ and the directly generated image $I^{\mrm{direct}}_{\mrm{half}}$ together. The result is called $I^{\mrm{blended}}_{\mrm{half}}$.
\end{itemize}
The half-resolution rotator outputs $I^{\mrm{flow}}_{\mrm{half}}$, $I^{\mrm{warped}}_{\mrm{half}}$, $I^{\mrm{direct}}_{\mrm{half}}$, and $I^{\mrm{blended}}_{\mrm{half}}$. These image are used for training. However, at test time, $I^{\mrm{flow}}_{\mrm{half}}$ and $I^{\mrm{blended}}_{\mrm{half}}$ are used by the next network, the editor.

\subsection{Editor} \label{sec:editor-architecture}

\begin{figure}
  \centering
  \includegraphics[width=16cm]{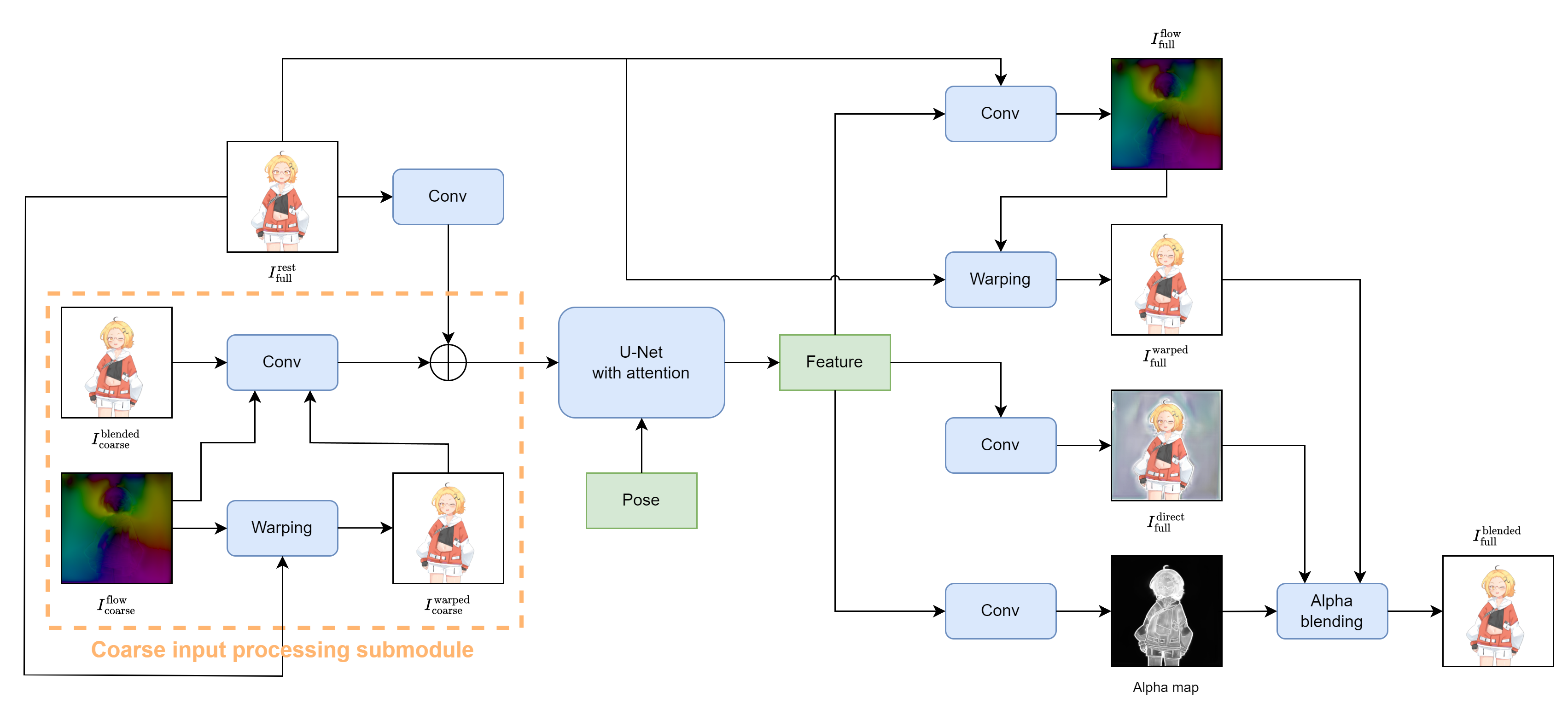}
  \caption{The new editor.}
  \label{fig:editor-architecture}
\end{figure}

The architecture of the editor is depicted in Figure~\ref{fig:editor-architecture}. It takes 4 inputs:
\begin{enumerate}
  \item $I^{\mrm{rest}}_{\mrm{full}}$, the original character image at the $512 \times 512$ resolution,
  \item $\mathbf{p}$, the 6-dimensional pose vector, 
  \item $I^{\mrm{blended}}_{\mrm{coarse}}$, which is $I^{\mrm{blended}}_{\mrm{half}}$ scaled up to $512 \times 512$, and 
  \item $I^{\mrm{flow}}_{\mrm{coarse}}$, which is $I^{\mrm{flow}}_{\mrm{half}}$ scaled up to $512 \times 512$. 
\end{enumerate}
The way the editor processes these input is quite similar to what the half-resolution rotator does. The pose vector is passed to the backbone directly. The input image $I^{\mrm{rest}}_{\mrm{full}}$ is convolved to create a $32 \times 512 \times 512$ tensor. The two other inputs are passed through what we call the ``coarse input processing submodule,'' which carries out the following steps.
\begin{itemize}
  \item First, the coarse appearance flow $I^{\mrm{flow}}_{\mrm{coarse}}$ is used to warped the original input image $I^{\mrm{rest}}_{\mrm{full}}$ to obtain the coarse warped image $I^{\mrm{warped}}_{\mrm{coarse}}$.
  \item Second, $I^{\mrm{blended}}_{\mrm{coarse}}$, $I^{\mrm{flow}}_{\mrm{coarse}}$, and $I^{\mrm{warped}}_{\mrm{coarse}}$ are concatenated, and the resulting tensor is convolved to form a $32 \times 512 \times 512$ tensor.
  \item Third, the result form the last step is added to the output of convolution layer that was applied to the original image to produce a $32 \times 512 \times 512$ tensor.
\end{itemize}
The resulting tensor is then passed to the backbone U-Net with attention. The output of the backbone is processed in the same way as what the half-resolution rotator does. This produces four tensors at full resolution: $I^{\mrm{flow}}_{\mrm{full}}$, $I^{\mrm{warped}}_{\mrm{full}}$, $I^{\mrm{direct}}_{\mrm{full}}$, and $I^{\mrm{blended}}_{\mrm{full}}$.

Note that, if we remove the coarse input processing submodule, the architecture of the editor would be exactly the same as the half-resolution rotator. Hence, the editor can be thought of as a network that also rotates the body given in the original image $I^{\mrm{rest}}_{\mrm{full}}$, but it takes the coarse inputs, $I^{\mrm{blended}}_{\mrm{coarse}}$ and $I^{\mrm{flow}}_{\mrm{coarse}}$, as hints. We will exploit this property in the training process of the editor.

\section{Full System's Training Details} \label{sec:training-details}

\subsection{Half-Resolution Rotator}

The half-resolution rotator is trained with the following 6-termed loss that is a combination of the L1 loss and the perceptual content loss \cite{Johnson:2016}. More concretely,
\begin{align*}
  \mcal{L}_{\mrm{HRR}} 
  &= \ell_{\mrm{L1}} \Big( \mathtt{L}_{\mrm{L1}}^{\mrm{warped}} 
  + \mathtt{L}_{\mrm{L1}}^{\mrm{direct}} 
  + \mathtt{L}_{\mrm{L1}}^{\mrm{blended}} \Big)
  + \ell_{\mrm{percept}} \Big( \mathtt{L}_{\mrm{percept}}^{\mrm{warped}}
  + \mathtt{L}_{\mrm{percept}}^{\mrm{direct}}
  + \mathtt{L}_{\mrm{percept}}^{\mrm{blended}} \Big).
\end{align*}
The $\ell_{\mrm{L1}}$ and $\ell_{\mrm{percept}}$ are loss weights, which change once during the training process. (More on this later.) The loss terms that have subscripts ``L1'' are given by \begin{align*}
  \mathtt{L}_{L1}^{\square} 
  &= \frac{\| I_{\mrm{half}}^{\square} - I_{\mrm{half}}^{\mrm{posed}} \|_1}{C \times H \times W}
\end{align*}
where $C$, $H$, and $W$ are the channels, height, and width of the tensors, respectively. The $I_{\mrm{half}}^{\mrm{posed}}$ is the groundtruth posed image in the training dataset scaled down to $256 \times 256$, and $I_{\mrm{half}}^{\square}$ are the outputs of the half-resolution rotator as defined in Section~\ref{sec:half-resolution-rotator-architecture}. The loss with subscripts ``percept'' is given by
\begin{align*}
  \mathtt{L}_{\mrm{percept}}^{\square} 
  &= \Phi(I_{\mrm{half}}^{\square} - I_{\mrm{half}}^{\mrm{posed}})
\end{align*}
and 
\begin{align*}
  \Phi(I_1, I_2) = \sum_{i=1}^3 c_i \big( \| \phi_i(I^{\mrm{rgb}}_1) - \phi_i(I^{\mrm{rgb}}_2) \|_1 + \| \phi_i(I^{\mrm{aaa}}_1) - \phi_i(I^{\mrm{aaa}}_2) \|_1 \big).
\end{align*}
Here, 
\begin{itemize}
  \item $\phi_i(\cdot)$ denote the feature tensor outputted by the $i$th used layer in the VGG16 network \cite{Simonyan:2015}, and we use the \texttt{relu1\_2}$, $\text{relu2\_2}, and \texttt{relu3\_3} layers.
  \item $c_i$ is the reciprocal of the number of components of $\phi_i(\cdot)$.
  \item $I^{\mrm{rgb}}$ denotes the 3-channel image formed by dropping the alpha channel of image $I$. 
  \item $I^{\mrm{aaa}}$ denotes the 3-channel image formed by repeating the alpha channel of $I$ three times. 
\end{itemize}
We compute the perceptual loss as two L1 loss terms because the VGG16 network accepts an RGB image as input whereas the images outputted by the half-resolution rotator have 4 channels. To speed up the computation of $\Phi(\cdot, \cdot)$, we evaluate it stochastically. We flip a coin with head probability of $3/4$. If it turns up head, we evaluate the term with $I^{\mrm{rgb}}$; otherwise, we evaluate the term with $I^{\mrm{aaa}}$. Of course, the terms are scaled with the reciprocal of the probability to make sure that the expectation has the right value.

Training is divided into two phases.
\begin{itemize}
  \item In the first phase, only the L1 losses are used. In other words, $\ell_{\mrm{L1}} = 1$, and $\ell_{\mrm{percept}} = 0$. The first phase lasts for 1 epoch (500K training examples).
  \item In the second phase, all loss terms are used. In particular, we set $\ell_{\mrm{L1}} = 1$, and $\ell_{\mrm{percept}} = 0.2$. The second phase lasts for 18 epochs (9M training examples).
\end{itemize}
We used the Adam optimizer with $\beta_1 = 0.5$ and $\beta_2 = 0.999$. The learning rate starts at $0$ and linearly increases to $10^{-4}$ over the first 5,000 training examples. The batch size was 16.

\subsection{Editor}

Training has three phases. In the first two phases, the coarse input processing submodule is dropped from the editor, making it temporarily a ``full-resolution rotator.'' The network is trained using the training process of the half-resolution rotator but now with the full resolution images instead of the half-resolution ones. 

In the third phase, we added the coarse input processing submodule back and train the network to minimize the following loss:
\begin{align*}
  \mcal{L}_{\mrm{ED}} = \lambda_{\mrm{L1}} \Big( \mcal{L}^{\mrm{warped}}_\mrm{L1} + \mcal{L}^{\mrm{direct}}_\mrm{L1} + \mcal{L}^{\mrm{blended}}_{\mrm{L1}} \Big) + \lambda_{\mrm{percept}}\Big( \mcal{L}_{\mrm{half}}^{\mrm{direct}} + \mcal{L}_{\mrm{half}}^{\mrm{blended}} + \mcal{L}_{\mrm{quad}}^{\mrm{direct}} + \mcal{L}_{\mrm{quad}}^{\mrm{blended}} \Big).
\end{align*}
We fixed $\lambda_{\mrm{L1}} = 1$ and $\lambda_{\mrm{percept}} = 0.2$. The L1 losses are given by
\begin{align*}
  \mcal{L}_{\mrm{L1}}^{\square} = \| I^{\square}_{\mrm{full}} - I^{\mrm{posed}}_{\mrm{full}} \|_1
\end{align*}
where $I^{\mrm{posed}}_{\mrm{full}}$ is the groundtruth posed image from the training dataset, and the $I^{\square}_{\mrm{full}}$ are the outputs of the editor as defined in Section~\ref{sec:editor-architecture}. The losses $\mcal{L}_{\mrm{half}}^{\mrm{direct}}$, $\mcal{L}_{\mrm{half}}^{\mrm{blended}}$, $\mcal{L}_{\mrm{quad}}^{\mrm{direct}}$, and $\mcal{L}_{\mrm{quad}}^{\mrm{blended}}$ are perceptual losses. The superscripts indicate the outputs of the editor that we compute the losses with, and the subscripts indicate how the losses are computed. The ``half'' subscript indicates that the images are scaled down to $256 \times 256$:
\begin{align*}
  \mcal{L}_{\mrm{half}}^{\square} = \Phi\Big( \textsc{Down}( I_{\mrm{full}}^{\square}), \textsc{Down}( I_{\mrm{full}}^{\mrm{posed}} ) \Big)
\end{align*}
where $\textsc{Down}(\cdot)$ denotes scaling a $512 \times 512$ image down to $256 \times 256$. The ``quad'' subscript indicates that the images are divided into four quadrants so that a $512 \times 512$ images becomes four $4 \times 256 \times 256$ images. The quadrants are then used to evaluate the perceptual losses.
\begin{align*}
  \mcal{L}^{\square}_{\mrm{quad}} = \sum_{i=1}^4 \Phi\Big( Q_i( I_{\mrm{full}}^{\square} ), Q_i( I_{\mrm{full}}^{\mrm{posed}} ) \Big)  
\end{align*}
where $Q_i(\cdot)$ extracts the $i$th quadrant from the argument. We found that evaluating the perceptual losses at $256 \times 256$ rather than $512 \times 512$ led to a network that produced sharper images.

Again, we use the same optimizers and learning rate schedule as those of the half-resolution rotator. The first phase lasts for 1 epoch (500K examples), the second 18 epochs (9M examples), and the third 18 epochs (9M examples).

\section{Computers Used for Speed Measurements} \label{sec:computers-for-speed-measurement}

We conducted experiments that measured time it took for the models to produce a single animation frame on the following 3 desktop computers.

\begin{itemize}
  \item {\bf Computer A} has two Nvidia Nvidia RTX A6000 GPUs, a 2.10 GHz Intel Xeon Silver 4310 CPU, and 128 GB of RAM. It represents a computer used primarily for machine learning research.
  \item {\bf Computer B} has an Nvidia Titan RTX GPU, a 3.60 GHz Intel Core i9-9900KF CPU, and 64 GB of RAM. It represents a high-end gaming PC.
  \item {\bf Computer C} has an Nvidia GeForce GTX 1080 Ti GPU, a 3.70 GHz Intel Core i7-8700K CPU, and 32 GB of RAM. It represents a typical (yet somewhat outdated) gaming PC.
\end{itemize}

\section{Student Face Morpher's Loss Function} \label{sec:student-training-details}

\begin{figure}
  \centering
  \begin{tabular}{cc}
    \frame{\includegraphics[width=4cm]{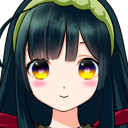}} & 
    \frame{\includegraphics[width=4cm]{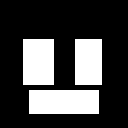}} \\
    (a) Character face & (b) Mask
  \end{tabular}
  \caption{Binary mask that covers movable facial parts of a character.}
  \label{fig:face-organ-mask}
\end{figure}

The student face morpher is trained to minimize the following loss:
\begin{align*}
  \mcal{L}_{\mrm{fm}} = E_{\ve{p} \sim p_{\mrm{pose}}} \Big[ \big\| S^{\mrm{fm}}(I_{\mrm{in}},\ve{p}) - T^{\mrm{fm}}(I_{\mrm{in}}, \ve{p}) \big\|_1  + \lambda_{\mrm{fm}} \big\| M \odot ( S^{\mrm{fm}}(I_{\mrm{in}},\ve{p}) - T^{\mrm{fm}}(I_{\mrm{in}}, \ve{p})) \big\|_1 \Big].
\end{align*}
Here, 
\begin{itemize}
  \item $I_{\mrm{in}}$ is the image of the character that we want to create a specialized student model of. 
  \item $\ve{p}$ is a pose vector, which is sampled from the training dataset of the full system.
  \item $p_{\mrm{pose}}$ is the uniform distribution over the poses in the training dataset. 
  \item $S^{fm}(\cdot, \cdot)$ is the student face morpher.
  \item $T^{fm}(\cdot, \cdot)$ is the teacher face morpher, which comes from the full system in Section~\ref{sec:baseline}.
  \item $M$ is a binary mask that covers all the movable facial organs: eyebrows, eyes, mouth, and chin. This mask has to be created for each individual character. See Figure~\ref{fig:face-organ-mask} for an example.
  \item Lastly, $\lambda_{fm}$ is a weighting constant, whose value is $20$ in all experiments.
\end{itemize}    

\end{document}